\documentclass[runningheads]{llncs}
\usepackage{graphicx}
\usepackage{amsmath,amssymb} 
\usepackage{color}
\usepackage{makecell}
\usepackage{times}
\usepackage{epsfig}
\usepackage{graphicx}
\usepackage{float}
\usepackage{wrapfig}
\usepackage{amsmath,amssymb,amsthm}
\usepackage{algorithm,algorithmicx,algpseudocode}
\usepackage{bm,xspace}
\usepackage{comment}
\usepackage{verbatim}
\usepackage{multirow}
\usepackage{balance}
\usepackage{url}
\usepackage{booktabs}
\usepackage{etoolbox,siunitx}
\usepackage{calc}
\usepackage{pifont,hologo}
\usepackage{color}
\usepackage{capt-of}

\def\bfn{\mathbf{n}}

\def\bfC{\mathbf{C}}

\def\bfF{\mathbf{F}}

\def\bfO{\mathbf{O}}
\def\bfP{\mathbf{P}}

\def\calE{\mathcal{E}}

\def\calM{\mathcal{M}}
\def\calN{\mathcal{N}}

\def\calV{\mathcal{V}}

\def\bbR{\mathbb{R}}

\def\etal{\emph{et al.}}
\def\eg{\emph{e.g.}}

\newcommand{\figref}[1]{Fig. \ref{#1}}
\newcommand{\secref}[1]{Sec. \ref{#1}}
\newcommand{\tabref}[1]{Tab. \ref{#1}}

\begin{document}
\title{Pixel2Mesh: Generating 3D Mesh Models \\from Single RGB Images} 
\makeatletter
\def\blfootnote{\xdef\@thefnmark{}\@footnotetext}
\makeatother
\titlerunning{Pixel2Mesh}

%
%


\author{Nanyang Wang$^{1\,\star}$, ~~Yinda Zhang$^{2\,\star}$, ~~Zhuwen Li$^{3\,\star}$, \\Yanwei Fu$^{4}$, ~~Wei Liu$^{5}$, ~~Yu-Gang Jiang$^{1\,\dag}$}
\authorrunning{N. Wang$^\star$, Y. Zhang$^\star$, Z. Li$^\star$, Y. Fu, W, Liu, Y. Jiang}

\institute{\textsuperscript{1}Shanghai Key Lab of Intelligent Information Processing,\\ School of Computer Science, Fudan University\\
\textsuperscript{2}Princeton University
~\textsuperscript{3}Intel Labs
~\textsuperscript{4}School of Data Science, Fudan University
~\textsuperscript{5}Tencent AI Lab\\
\email{nywang16@fudan.edu.cn}
~~~\email{yindaz@cs.princeton.edu}
~~~\email{lzhuwen@gmail.com}\\
\email{yanweifu@fudan.edu.cn}
~~~\email{wl2223@columbia.edu}
~~~\email{ygj@fudan.edu.cn}}
\blfootnote{$^\star$ indicates equal contributions.}
\blfootnote{$^\dag$ indicates corresponding author.}
\maketitle              
\begin{abstract}
We propose an end-to-end deep learning architecture that produces a 3D shape in triangular mesh from a single color image.
Limited by the nature of deep neural network, previous methods usually represent a 3D shape in volume or point cloud, and it is non-trivial to convert them to the more ready-to-use mesh model.
Unlike the existing methods, our network represents 3D mesh in a graph-based convolutional neural network and produces correct geometry by progressively deforming an ellipsoid, leveraging perceptual features extracted from the input image.
We adopt a coarse-to-fine strategy to make the whole deformation procedure stable, and define various of mesh related losses to capture properties of different levels to guarantee visually appealing and physically accurate 3D geometry.
Extensive experiments show that our method not only qualitatively produces mesh model with better details, but also achieves higher 3D shape estimation accuracy compared to the state-of-the-art.

\keywords{3D shape generation \and Graph convolutional neural network \and Mesh reconstruction \and Coarse-to-fine \and End-to-end framework}
\end{abstract}

\section{Introduction}
\label{sec:intro}

\begin{figure}[t]
\centering
\includegraphics[width=1\linewidth]{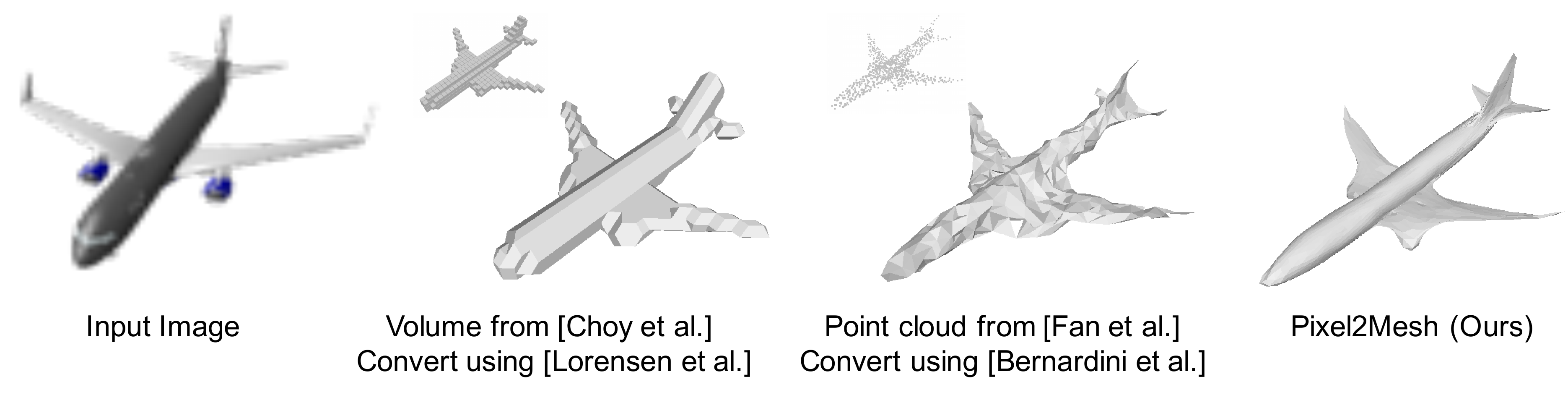}
\caption{Given a single color image and an initial mesh, our method can produce a high-quality mesh that contains details from the example.}
\label{fig:teaser}
\end{figure}



Inferring 3D shape from a single perspective is a fundamental human vision functionality but is extremely challenging for computer vision.
Recently, great success has been achieved for 3d shape generation from a single color image using deep learning techniques \cite{ChoyXGCS16,FanSG16}.
Taking advantage of convolutional layers on regular grids or multi-layer perception, the estimated 3D shape, as the output of the neural network, is represented as either a volume \cite{ChoyXGCS16} or point cloud \cite{FanSG16}. 
However, both representations lose important surface details, and is non-trivial to reconstruct a surface model (\figref{fig:teaser}), i.e. a mesh, which is more desirable for many real applications since it is lightweight, capable of modelling shape details, easy to deform for animation, to name a few. 

In this paper, we push along the direction of single image reconstruction, and propose an algorithm to extract a 3D triangular mesh from a single color image. Rather than directly synthesizing, our model learns to deform a mesh from a mean shape to the target geometry.
This benefits us from several aspects. First, deep network is better at predicting residual, e.g. a spatial deformation, rather than structured output, e.g. a graph. Second, a series of deformations can be added up together, which allows shape to be gradually refined in detail. It also enables the control of the trade-off between the complexity of the deep learning model and the quality of the result.
Lastly, it provides the chance to encode any prior knowledge to the initial mesh, e.g. topology. As a pioneer study, in this work, we specifically work on objects that can be approximated using 3D mesh with genus 0 by deforming an ellipsoid with a fixed size. In practice, we found most of the commonly seen categories can be handled well under this setting, e.g. car, plane, table, etc. To achieve this goal, there are several inherent challenges.

The first challenge is how to represent a mesh model, which is essentially an irregular graph, in a neural network and still be capable of extracting shape details effectively from a given color image represented in a 2D regular grid.
It requires the integration of the knowledge learned from two data modalities.
On the 3D geometry side, we directly build a graph based fully convolutional network (GCN) \cite{BronsteinBLSV17,DefferrardBV16,KipfW16} on the mesh model, where the vertices and edges in the mesh are directly represented as nodes and connections in a graph.
Network feature encoding information for 3D shape is saved on each vertex.
Through forward propagation, the convolutional layers enable feature exchanging across neighboring nodes, and eventually regress the 3D location for each vertex.
On the 2D image side, we use a VGG-16 like architecture to extract features as it has been demonstrated to be successful for many tasks \cite{GatysEB16,LiCK18}.
To bridge these two, we design a perceptual feature pooling layer which allows each node in the GCN to pool image features from its 2D projection on the image, which can be readily obtained by assuming known camera intrinsic matrix.
The perceptual feature pooling is enabled once after several convolutions (i.e. a deformation block described in \secref{sec:cmrn}) using updated 3D locations, and hence the image features from correct locations can be effectively integrated with 3D shapes.

Given the graph representation, the next challenge is how to update the vertex location effectively towards ground truth. 
In practice, we observe that network trained to directly predict mesh with a large number of vertices is likely to make mistake in the beginning and hard to fix later.
One reason is that a vertex cannot effectively retrieve features from other vertices with a number of edges away, i.e. the limited receptive field.
To solve this problem, we design a graph unpooling layer, which allows the network to initiate with a smaller number of vertices and increase during the forward propagation.
With fewer vertices at the beginning stages, the network learns to distribute the vertices around to the most representative location, and then add local details as the number of vertices increases later.
Besides the graph unpooling layer, we use a deep GCN enhanced by shortcut connections \cite{HeZRS16} as the backbone of our architecture, which enables large receptive fields for global context and more steps of movements.

Representing the shape in graph also benefits the learning procedure.
The known connectivity allows us to define higher order loss functions across neighboring nodes, which are important to regularize 3D shapes.
Specifically, we define a surface normal loss to favor smooth surface; an edge loss to encourage uniform distribution of mesh vertices for high recall; and a laplacian loss to prevent mesh faces from intersecting each other.
All of these losses are essential to generate quality appealing mesh model, and none of them can be trivially defined without the graph representation.

The contributions of this paper are mainly in three aspects.
First, we propose a novel end-to-end neural network architecture that generates a 3D mesh model from a single RGB image.
Second, we design a projection layer which incorporates perceptual image features into the 3D geometry represented by GCN.
Third, our network predict 3D geometry in a coarse to fine fashion, which is more reliable and easy to learn.

\section{Related Work}
\label{sec:related}

3D reconstruction has been well studied based on the multi-view geometry (MVG) \cite{Multiview04} in the literature. The major research directions include structure from motion (SfM) \cite{SchonbergerF16} for large-scale high-quality reconstruction and simultaneous localization and mapping (SLAM) \cite{CadenaCCLSNRL16} for navigation. Though they are very successful in these scenarios, they are restricted by 1) the coverage that the multiple views can give and 2) the appearance of the object that wants to reconstruct. The former restriction means MVG cannot reconstruct unseen parts of the object, and thus it usually takes a long time to get enough views for a good reconstruction; the latter restriction means MVG cannot reconstruct non-lambertian (e.g. reflective or transparent) or textureless objects. These restrictions lead to the trend of resorting to learning based approaches.

Learning based approaches usually consider single or few images, as it largely relies on the shape priors that it can learn from data. Early works can be traced back to Hoiem \etal \cite{HoiemEH07} and Saxena \etal \cite{SaxenaSN09}. Most recently, with the success of deep learning architectures and the release of large-scale 3D shape datasets such as ShapeNet \cite{ChangFGHHLSSSSX15}, learning based approaches have achieved great progress. Huang \etal \cite{HuangWK15} and Su \etal \cite{SuHMLG14} retrieve shape components from a large dataset, assemble them and deform the assembled shape to fit the observed image. However, shape retrieval from images itself is an ill-posed problem. To avoid this problem, Kar \etal \cite{KarTCM15} learns a 3D deformable model for each object category and capture the shape variations in different images. However, the reconstruction is limited to the popular categories and its reconstruction result is usually lack of details.
Another line of research is to directly learn 3D shapes from single images. Restricted by the prevalent grid-based deep learning architectures, most works \cite{ChoyXGCS16,GirdharFRG16} outputs 3D voxels, which are usually with low resolutions due to the memory constraint on a modern GPU. Most recently, Tatarchenko \etal \cite{TatarchenkoDB17} have proposed an octree representation, which allows to reconstructing higher resolution outputs with a limited memory budget. However, a 3D voxel is still not a popular shape representation in game and movie industries.
To avoid drawbacks of the voxel representation, Fan \etal \cite{FanSG16} propose to generate point clouds from single images. The point cloud representation has no local connections between points, and thus the point positions have a very large degree of freedom. Consequently, the generated point cloud is usually not close to a surface and cannot be used to recover a 3D mesh directly.
Besides these typical 3D representations, there is an interesting work \cite{SinhaUHR17} which uses a so-called ``geometry image'' to represent a 3D shape. Thus, their network is a 2D convolutional neural network which conducts an image to image mapping. Our works are mostly related to the two recent works \cite{KatoUH2018} and \cite{PontesKSLF2017}. However, the former adopts simple silhouette supervision, and hence does not perform well for complicated objects such as car, lamp, etc; the latter needs a large model repository to generate a combined model.

Our base network is a graph neural network \cite{ScarselliGTHM09}; this architecture has been adopted for shape analysis \cite{YiSGG16}. In the meanwhile, there are charting-based methods which directly apply convolutions on surface manifolds \cite{BoscainiMRB16,MasciBBV15,MontiBMRSB17} for shape analysis.
As far as we know, these architectures have never been adopted for 3D reconstruction from single images, though graph and surface manifold are natural representations for meshed objects. For a comprehensive understanding of the graph neural network, the charting-based methods and their applications, please refer to this survey \cite{BronsteinBLSV17}.

\section{Method}

\subsection{Preliminary: Graph-based Convolution}
We first provide some background about graph based convolution; more detailed introduction can be found in \cite{BronsteinBLSV17}.
A 3D mesh is a collection of vertices, edges and faces that defines the shape of a 3D object; it can be represented by a graph $\calM=(\calV, \calE, \bfF)$, where $\calV=\{v_i\}^{N}_{i=1}$ is the set of $N$ vertices in the mesh, $\calE=\{e_i\}^{E}_{i=1}$ is the set of $E$ edges with each connecting two vertices, and $\bfF=\{f_i\}^{N}_{i=1}$ are the feature vectors attached on vertices.
A graph based convolutional layer is defined on irregular graph as: 
\begin{equation}
f_p^{l+1}= w_0 f_p^l + \sum_{q\in \calN(p)} w_1 f_q^l
\label{eqn:gcn}
\end{equation}
where $f_p^l\in\bbR^{d_l}, f_p^{l+1}\in\bbR^{d_{l+1}}$ are the feature vectors on vertex $p$ before and after the convolution, and $\calN(p)$ is the neighboring vertices of $p$; $w_0$ and $w_1$ are the learnable parameter matrices of $d_l\times d_{l+1}$ that are applied to all vertices. Note that $w_1$ is shared for all edges, and thus \eqref{eqn:gcn} works on nodes with different vertex degrees.
In our case, the attached feature vector $f_p$ is the concatenation of the 3D vertex coordinate, feature encoding 3D shape, and feature learned from the input color image (if they exist).
Running convolutions updates the features, which is equivalent as applying a deformation.

\subsection{System Overview}
\label{sec:overiew}
\begin{figure}[t]
\centering
\includegraphics[width=1\linewidth]{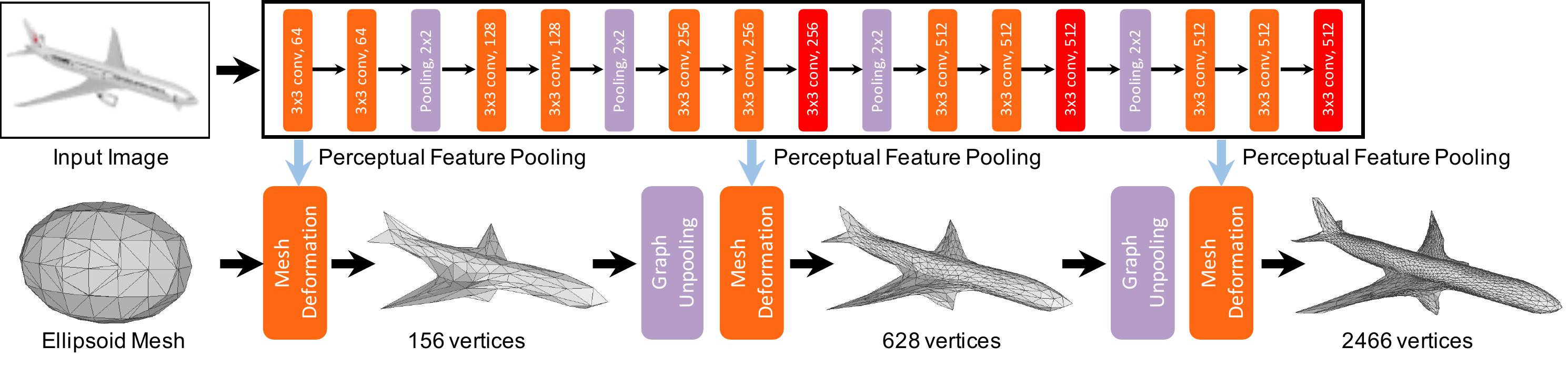}
\caption{The cascaded mesh deformation network. Our full model contains three mesh deformation blocks in a row. Each block increases mesh resolution and estimates vertex locations, which are then used to extract perceptual image features from the 2D CNN for the next block.}
\label{fig:cmrn}
\end{figure}

Our model is an end-to-end deep learning framework that takes a single color image as input and produces a 3D mesh model in camera coordinate. The overview of our framework is illustrated in \figref{fig:cmrn}. 
The whole network consists an image feature network and a cascaded mesh deformation network.
The image feature network is a 2D CNN that extract perceptual feature from the input image, which is leveraged by the mesh deformation network to progressively deform an ellipsoid mesh into the desired 3D model.
The cascaded mesh deformation network is a graph-based convolution network (GCN), which contains three deformation blocks intersected by two graph unpooling layers.
Each deformation block takes an input graph representing the current mesh model with the 3D shape feature attached on vertices, and produces new vertices locations and features.
Whereas the graph unpooling layers increase the number of vertices to increase the capacity of handling details, while still maintain the triangular mesh topology.
Starting from a smaller number of vertices, our model learns to gradually deform and add details to the mesh model in a coarse-to-fine fashion.
In order to train the network to produce stable deformation and generate an accurate mesh, we extend the Chamfer Distance loss used by Fan \etal \cite{FanSG16} with three other mesh specific loss -- Surface normal loss, Laplacian regularization loss, and Edge length loss.
The remaining part of this section describes details of these components.

\subsection{Initial ellipsoid}
\label{sec:init}
Our model does not require any prior knowledge of the 3D shape, and always deform from an initial ellipsoid with average size placed at the common location in the camera coordinate.
The ellipsoid is centered at 0.8m in front of the camera with 0.2m, 0.2m, 0.4m as the radius of three axis.
The mesh model is generated by implicit surface algorithm in Meshlab \cite{meshlab} and contains 156 vertices.
We use this ellipsoid to initialize our input graph, where the initial feature contains only the 3D coordinate of each vertex.

\subsection{Mesh deformation block}
\label{sec:cmrn}
\begin{figure}[t]
\centering
\includegraphics[width=1\linewidth]{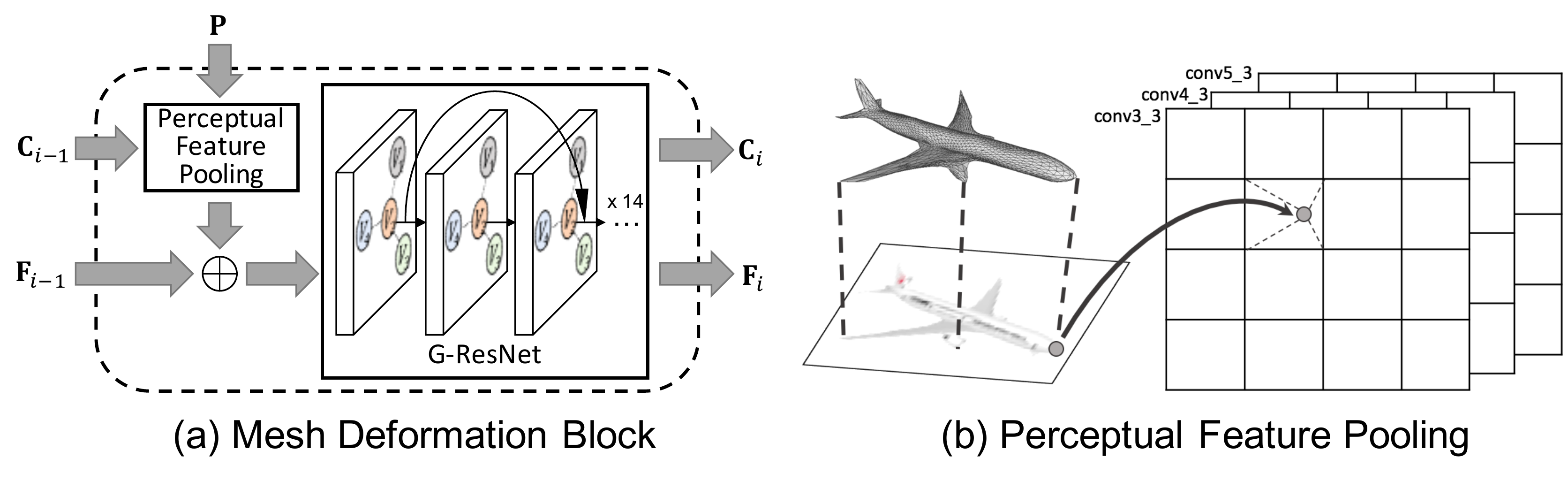}
\caption{(a) The vertex locations $\bfC_i$ are used to extract image features, which are then combined with vertex features $\bfF_i$ and fed into G-ResNet. $\bigoplus$ means a concatenation of the features. (b) The 3D vertices are projected to the image plane using camera intrinsics, and perceptual feature is pooled from the 2D-CNN layers using bilinear interpolation.}
\label{fig:block}
\end{figure}

The architecture of mesh deformation block is shown in \figref{fig:block} (a).
In order to generate 3D mesh model that is consistent with the object shown in the input image, the deformation block need to pool feature ($\bfP$) from the input image. This is done in conjunction with the image feature network and a perceptual feature pooling layer given the location of vertex ($\bfC_{i-1}$) in the current mesh model.
The pooled perceptual feature is then concatenated with the 3D shape feature attached on the vertex from the input graph ($\bfF_{i-1}$) and fed into a series of graph based ResNet (G-ResNet).
The G-ResNet produces, also as the output of the mesh deformation block, the new coordinates ($\bfC_i$) and 3d shape feature ($\bfF_i$) for each vertex.

\subsubsection{Perceptual feature pooling layer}
We use a VGG-16 architecture up to layer conv5\_3 as the image feature network as it has been widely used.
Given the 3D coordinate of a vertex, we calculate its 2D projection on input image plane using camera intrinsics, and then pool the feature from four nearby pixels using bilinear interpolation. In particular, we concatenate feature extracted from layer `conv3\_3', `conv4\_3', and `conv5\_3', which results in a total dimension of 1280.
This perceptual feature is then concatenated with the 128-dim 3D feature from the input mesh, which results in a total dimension of 1408. This is illustrated in \figref{fig:block} (b). Note that in the first block, the perceptual feature is concatenated with the 3-dim feature (coordinate) since there is no learnt shape feature at the beginning.


\subsubsection{G-ResNet}
After obtaining 1408-dim feature for each vertex representing both 3D shape and 2D image information, we design a graph based convolutional neural network to predict new location and 3D shape feature for each vertex.
This requires efficient exchange of the information between vertices.
However, as defined in \eqref{eqn:gcn}, each convolution only enables the feature exchanging between neighboring pixels, which severely impairs the efficiency of information exchanging. This is equivalent as the small receptive field issue on 2D CNN.

To solve this issue, we make a very deep network with shortcut connections \cite{HeZRS16} and denote it as G-ResNet (\figref{fig:block} (a)). In this work, the G-ResNet in all blocks has the same structure, which consists of 14 graph residual convolutional layers with 128 channels. The serial of G-ResNet block produces a new 128-dim 3D feature. In addition to the feature output, there is a branch which applies an extra graph convolutional layer to the last layer features and outputs the 3D coordinates of the vertex.


\subsection{Graph unpooling layer}
\begin{figure}[t]
\centering
\includegraphics[width=\linewidth]{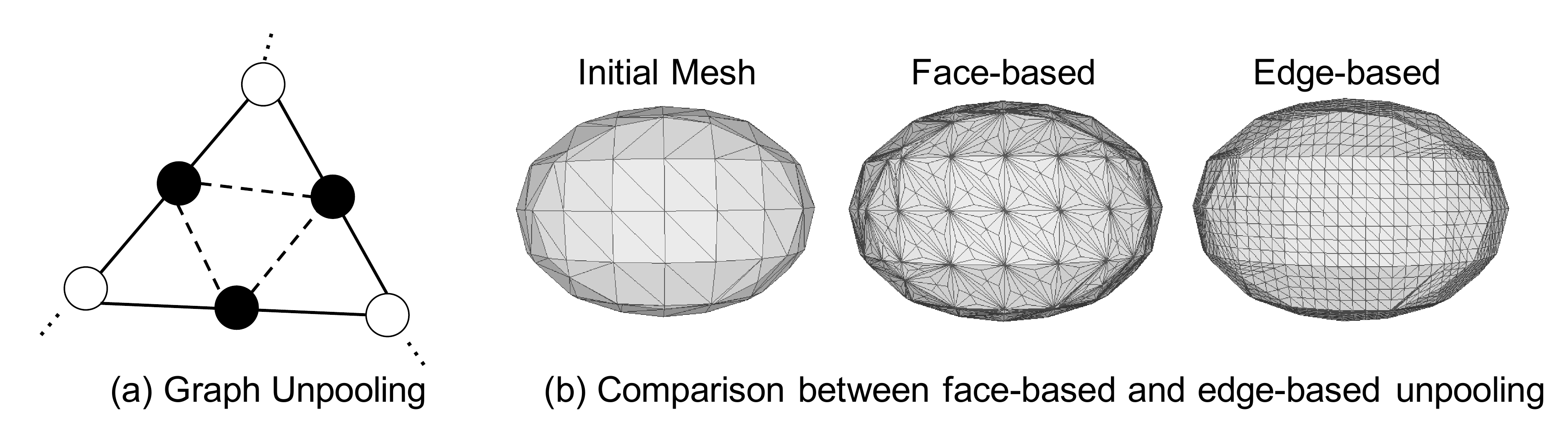}
\caption{(a) Black vertices and dashed edges are added in the unpooling layer. (b) The face based unpooling leads to imbalanced vertex degrees, while the edge-based unpooling remains regular.}
\label{fig:unpooling}
\end{figure}

The goal of unpooling layer is to increase the number of vertex in the GCNN.
It allows us to start from a mesh with fewer vertices and add more only when necessary, which reduces memory costs and produces better results.
A straightforward approach is to add one vertex in the center of each triangle and connect it with the three vertices of the triangle (\figref{fig:unpooling} (b) Face-based). However, this causes imbalanced vertex degrees, i.e. number of edges on vertex.
Inspired by the vertex adding strategy of the mesh subdivision algorithm prevalent in computer graphics, we add a vertex at the center of each edge and connect it with the two end-point of this edge (\figref{fig:unpooling} (a)). 
The 3D feature for newly added vertex is set as the average of its two neighbors.
We also connect three vertices if they are added on the same triangle (dashed line.)
Consequently, we create 4 new triangles for each triangle in the original mesh, and the number of vertex is increased by the number of edges in the original mesh.
This edge-based unpooling uniformly upsamples the vertices as shown in \figref{fig:unpooling} (b) Edge-based.

\subsection{Losses}
We define four kinds of losses to constrain the property of the output shape and the deformation procedure to guarantee appealing results.
We adopt the Chamfer loss \cite{FanSG16} to constrain the location of mesh vertices, a normal loss to enforce the consistency of surface normal, a laplacian regularization to maintain relative location between neighboring vertices during deformation, and an edge length regularization to prevent outliers. 
These losses are applied with equal weight on both the intermediate and final mesh.

Unless otherwise stated, we use $p$ for a vertex in the predicted mesh, $q$ for a vertex in the ground truth mesh, $\calN(p)$ for the neighboring pixel of $p$, till the end of this section.


\subsubsection{Chamfer loss}
The Chamfer distance measures the distance of each point to the other set:
$l_c = \sum_{p}\min_{q}\|p-q\|_2^2 + \sum_{q}\min_{p}\|p-q\|_2^2.$
It is reasonably good to regress the vertices close to its correct position, however is not sufficient to produce nice 3D mesh (see the result of Fan et al. \cite{FanSG16} in \figref{fig:teaser}).

\subsubsection{Normal loss}
We further define loss on surface normal to characterize high order properties:
$l_n=\sum_{p}\sum_{\substack{q=\mathrm{arg}\min_q(\|p-q\|_2^2)}}\|\langle p-k,\bfn_q\rangle\|_2^2, ~\mathrm{s.t}.~ k\in \calN(p)$
where $q$ is the closest vertex for $p$ that is found when calculating the chamfer loss, $k$ is the neighboring pixel of $p$, $\langle\cdot,\cdot\rangle$ is the inner product of two vectors, and $\bfn_q$ is the observed surface normal from ground truth.

Essentially, this loss requires the edge between a vertex with its neighbors to perpendicular to the observation from the ground truth. 
One may find that this loss does not equal to zero unless on a planar surface.
However, optimizing this loss is equivalent as forcing the normal of a locally fitted tangent plane to be consistent with the observation, which works practically well in our experiment.
Moreover, this normal loss is fully differentiable and easy to optimize.


\subsubsection{Regularization}
Even with the Chamfer loss and Normal loss, the optimization is easily stucked in some local minimum. More specifically, the network may generate some super large deformation to favor some local consistency, which is especially harmful at the beginning when the estimation is far from ground truth, and causes flying vertices (\figref{fig:ablation}).

\paragraph{Laplacian regularization} To handle these problem, we first propose a Laplacian term to prevent the vertices from moving too freely, which potentially avoids mesh self-intersection. The laplaician term serves as a local detail preserving operator, that encourages neighboring vertices to have the same movement. 
In the first deformation block, it acts like a surface smoothness term since the input to this block is a smooth-everywhere ellipsoid; starting from the second block, it prevents the 3D mesh model from deforming too much, so that only fine-grained details are added to the mesh model.
To calculate this loss, we first define a laplacian coordinate for each vertex $p$ as
$\delta_p=p-\sum_{k\in \calN(p)} \frac{1}{\|\calN(p)\|}k,$
and the laplacian regularization is defined as:
$l_{lap}=\sum_{p}\|\delta^\prime_p-\delta_p\|_2^2,$
where $\delta^\prime_p$ and $\delta_p$ are the laplacian coordinate of a vertex after and before a deformation block.

\paragraph{Edge length regularization.} To penalize flying vertices, which ususally cause long edge, we add an edge length regularization loss:
$l_{loc}=\sum_{p}\sum_{k\in \calN(p)}\|p-k\|_2^2.$

The overall loss is a weighted sum of all four losses, $l_{all} = l_c+\lambda_1l_n+\lambda_2l_{lap}+\lambda_3l_{loc}$, where $\lambda_1=1.6e-4$, $\lambda_2=0.3$ and $\lambda_3=0.1$ are the hyperparameters which balance the losses and fixed for all the experiments.

\section{Experiment}
\label{sec:exp}

In this section, we perform an extensive evaluation on our model. In addition to comparing with previous 3D shape generation works for evaluating the reconstruction accuracy, we also analyse the importance of each component in our model. Qualitative results on both synthetic and real-world images further show that our model produces triangular meshes with smooth surfaces and still maintains details depicted in the input images.

\subsection{Experimental setup}
\subsubsection{Data.}
We use the dataset provided by Choy et al. \cite{ChoyXGCS16}. The dataset contains rendering images of 50k models belonging to 13 object categories from ShapeNet \cite{ChangFGHHLSSSSX15}, which is a collection of 3D CAD models that are organized according to the WordNet hierarchy.
A model is rendered from various camera viewpoints, and camera intrinsic and extrinsic matrices are recorded.
For fair comparison, we use the same training/testing split as in Choy et. al. \cite{ChoyXGCS16}.


\subsubsection{Evaluation Metric.} 
We adopt the standard 3D reconstruction metric.
We first uniformly sample points from our result and ground truth. We calculate precision and recall by checking the percentage of points in prediction or ground truth that can find a nearest neighbor from the other within certain threshold $\tau$. A F-score \cite{KnapitschPZK17} as the harmonic mean of precision and recall is then calculated.
Following Fan et. al. \cite{FanSG16}, we also report the Chamfer Distance (CD) and Earth Mover's Distance (EMD). For F-Score, larger is better. For CD and EMD, smaller is better.

On the other hand, we realize that the commonly used evaluation metrics for shape generation may not thoroughly reflect the shape quality. They often capture occupancy or point-wise distance rather than surface properties, such as continuity, smoothness, high-order details, for which a standard evaluation metric is barely missing in literature. Thus, we recommend to pay attention on qualitative results for better understanding of these aspects.

\subsubsection{Baselines.} We compare the presented approach to the most recent single image reconstruction approaches. Specifically, we compare with two state-of-the-art methods - Choy et. al. \cite{ChoyXGCS16} (3D-R2N2) producing 3D volume, and Fan et. al. \cite{FanSG16} (PSG) producing point cloud.
Since the metrics are defined on point cloud, we can evaluate PSG directly on its output, our method by uniformly sampling point on surface, and 3D-R2N2 by uniformly sampling point from mesh created using the Marching Cube \cite{LorensenC87} method.

We also compare to Neural 3D Mesh Renderer (N3MR) \cite{KatoUH2018} which is so far the only deep learning based mesh generation model with code public available. For fair comparison, the models are trained with the same data using the same amount of time.

\subsubsection{Training and Runtime.} Our network receives input images of size $224\times 224$, and initial ellipsoid with 156 vertices and 462 edges. The network is implemented in Tensorflow and optimized using Adam with weight decay 1e-5. The batch size is 1; the total number of training epoch is 50; the learning rate is initialized as 3e-5 and drops to 1e-5 after 40 epochs. The total training time is 72 hours on a Nvidia Titan X. During testing, our model takes 15.58ms to generate a mesh with 2466 vertices.

\subsection{Comparison to state of the art}
\begin{table}[t]
\centering
\setlength{\tabcolsep}{0.95mm}
\renewcommand{\arraystretch}{1.2}
\small
\begin{tabular}{@{}lccccccccc@{}}
\hline
Threshold & \multicolumn{4}{@{}c@{}}{$\tau$} & \multicolumn{4}{@{}c@{}}{$2\tau$} \\
  \cmidrule(lr){2-5} \cmidrule(lr){6-9}
Category  & 3D-R2N2 & PSG & N3MR & Ours & 3D-R2N2 & PSG & N3MR & Ours\\
\hline
\hline
  plane & 41.46 & 68.20 & 62.10 & \bfseries71.12 & 63.23 & 81.22 & 77.15 & \bfseries81.38\\
  bench & 34.09 & 49.29 & 35.84 & \bfseries57.57 & 48.89 & 69.17 & 49.58 & \bfseries71.86\\
  cabinet & 49.88 & 39.93 & 21.04 & \bfseries60.39 & 64.83 & 67.03 & 35.16 & \bfseries77.19\\
  car & 37.80 & 50.70 & 36.66 & \bfseries67.86 & 54.84 & 77.79 & 53.93 & \bfseries84.15\\
  chair & 40.22 & 41.60 & 30.25 & \bfseries54.38 & 55.20 & 63.70 & 44.59 & \bfseries70.42\\
  monitor & 34.38 & 40.53 & 28.77 & \bfseries51.39 & 48.23 & 63.64 & 42.76 & \bfseries67.01\\
  lamp & 32.35 & 41.40 & 27.97 & \bfseries48.15 & 44.37 & 58.84 & 39.41 & \bfseries61.50\\
  speaker & 45.30 & 32.61 & 19.46 & \bfseries48.84 & 57.86 & 56.79 & 32.20 & \bfseries65.61\\
  firearm & 28.34 & 69.96 & 52.22 & \bfseries73.20 & 46.87 & 82.65 & 63.28 & \bfseries83.47\\
  couch & 40.01 & 36.59 & 25.04 & \bfseries51.90 & 53.42 & 62.95 & 39.90 & \bfseries69.83\\
  table & 43.79 & 53.44 & 28.40 & \bfseries66.30 & 59.49 & 73.10 & 41.73 & \bfseries79.20\\
  cellphone & 42.31 & 55.95 & 27.96 & \bfseries70.24 & 60.88 & 79.63 & 41.83 & \bfseries82.86\\
  watercraft & 37.10 & 51.28 & 43.71 & \bfseries55.12 & 52.19 & \bfseries70.63 & 58.85 & 69.99\\
\hline
  mean & 39.01 & 48.58 & 33.80 & \bfseries59.72 & 54.62 & 69.78 & 47.72 & \bfseries74.19\\
\hline
\end{tabular}
\caption{F-score (\%) on the ShapeNet test set at different thresholds, where $\tau=10^{-4}$. Larger is better. Best results under each threshold are bolded.}
\label{tab:res_shapenet_f}
\end{table}

\begin{table}[t]
\centering
\setlength{\tabcolsep}{0.95mm}
\renewcommand{\arraystretch}{1.2}
\small
\begin{tabular}{@{}lccccccccccccc@{}}
\hline
\multirow{2}{*}{Category} & \multicolumn{4}{@{}c@{}}{CD} & \multicolumn{4}{@{}c@{}}{EMD}\\
  \cmidrule(lr){2-5} \cmidrule(lr){6-9}
  & 3D-R2N2 & PSG & N3MR & Ours & 3D-R2N2 & PSG & N3MR & Ours\\
\hline
\hline
  plane & 0.895 & \bfseries0.430 & 0.450 & 0.477 & 0.606 & \bfseries0.396 & 7.498 & 0.579\\
  bench & 1.891 & 0.629 & 2.268 & \bfseries0.624 & 1.136 & 1.113 & 11.766 & \bfseries0.965\\
  cabinet & 0.735 & 0.439 & 2.555 & \bfseries0.381 & \bfseries2.520 & 2.986 & 17.062 & 2.563\\
  car & 0.845 & 0.333 & 2.298 & \bfseries0.268 & 1.670 & 1.747 & 11.641 & \bfseries1.297\\
  chair & 1.432 & 0.645 & 2.084 & \bfseries0.610 & 1.466 & 1.946 & 11.809 & \bfseries1.399\\
  monitor & 1.707 & \bfseries0.722 & 3.111 & 0.755 & 1.667 & 1.891 & 14.097 & \bfseries1.536\\
  lamp & 4.009 & \bfseries1.193 & 3.013 & 1.295 & 1.424 & 1.222 & 14.741 & \bfseries1.314\\
  speaker & 1.507 & 0.756 & 3.343 & \bfseries0.739 & \bfseries2.732 & 3.490 & 16.720 & 2.951\\
  firearm & 0.993 & \bfseries0.423 & 2.641 & 0.453 & 0.688 & \bfseries0.397 & 11.889 & 0.667\\
  couch & 1.135 & 0.549 & 3.512 & \bfseries0.490 & 2.114 & 2.207 & 14.876 & \bfseries1.642\\
  table & 1.116 & 0.517 & 2.383 & \bfseries0.498 & 1.641 & 2.121 & 12.842 & \bfseries1.480\\
  cellphone & 1.137 & 0.438 & 4.366 & \bfseries0.421 & 0.912 & 1.019 & 17.649 & \bfseries0.724\\
  watercraft & 1.215 & \bfseries0.633 & 2.154 & 0.670 & 0.935 & 0.945 & 11.425 & \bfseries0.814\\
\hline
  mean & 1.445 & 0.593 & 2.629 & \bfseries0.591 & 1.501 & 1.653 & 13.386 & \bfseries1.380\\
\hline
\end{tabular}
\caption{CD and EMD on the ShapeNet test set. Smaller is better. Best results under each threshold are bolded.}
\label{tab:res_shapenet_cd}
\end{table}

\tabref{tab:res_shapenet_f} shows the F-score with different thresholds of different methods. Our approach outperforms the other methods in all categories except watercraft. Notably, our results are significantly better than the others in all categories under a smaller threshold $\tau$, showing at least 10\% F-score improvement.
N3MR does not perform well, and its result is about 50\% worse than ours, probably because their model only learns from limited silhouette signal in images and lacks of explicit handling of the 3D mesh.

\begin{figure}[t]
\centering
\includegraphics[width=1\linewidth]{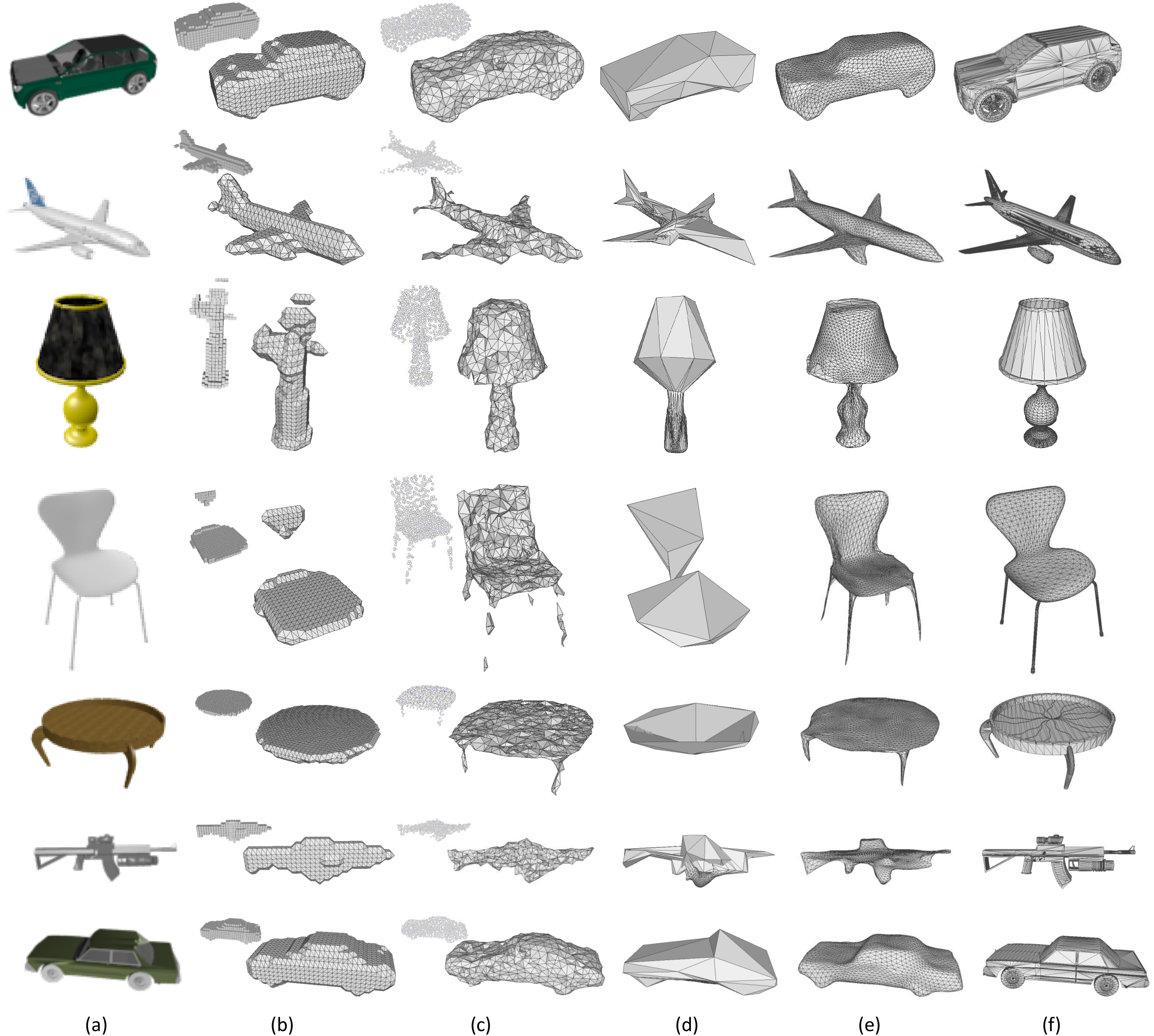}
\caption{Qualitative results. (a) Input image; (b) Volume from 3D-R2N2 \cite{ChoyXGCS16}, converted using Marching Cube \cite{LorensenC87}; (c) Point cloud from PSG \cite{FanSG16}, converted using ball pivoting \cite{BernardiniMRST99}; (d) N3MR\cite{KatoUH2018}; (e) Ours; (f) Ground truth.}
\label{fig:qual_res}
\end{figure}

We also show the CD and EMD for all categories in \tabref{tab:res_shapenet_cd}. Our approach outperforms the other methods in most categories and achieves the best mean score. The major competitor is PSG, which produces a point cloud and has the most freedom; this freedom leads to smaller CD and EMD, however does not necessarily leads to a better mesh model without proper regularization. To demonstrate this, we show the qualitative results to analyze why our approach outperforms the others.
\figref{fig:qual_res} shows the visual results. To compare the quality of mesh model, we convert volumetric and point cloud to mesh using standard approaches \cite{LorensenC87,BernardiniMRST99}. As we can see, the 3D volume results produced by 3D-R2N2 lack of details due to the low resolution, \eg, the legs are missing in the chair example as shown in the 4-th row of \figref{fig:qual_res}. We tried octree based solution \cite{TatarchenkoDB17} to increase the volume resolution, but found it still hard to recover surface level details as much as our model. PSG produces sparse 3D point clouds, and it is non-trivial to recover meshes from them.
This is due to the applied Chamfer loss acting like a regression loss which gives too much degree of freedom to the point cloud. N3MR produces very rough shape, which might be sufficient for some rendering tasks, however cannot recover complicated objects such as chairs and tables. 
In contrast, our model does not suffer from these issues by leveraging a mesh representation, integration of perceptual feature, and carefully defined losses during the training. Our result is not restricted by the resolution due to the limited memory budget and contains both smooth continuous surface and local details.

\subsection{Ablation Study} 
\begin{figure}[tbhp]
\centering
\includegraphics[width=1\linewidth]{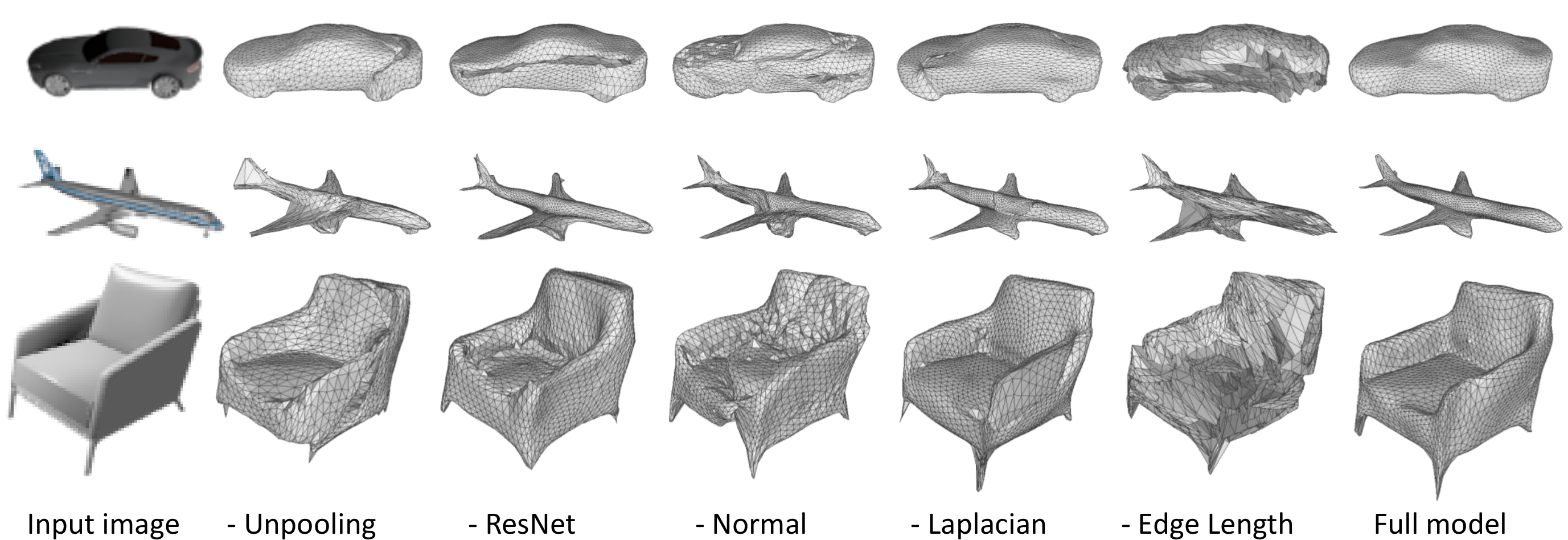}
\caption{Qualitative results for ablation study. This figure truly reflects the contribution of each components especially for the regularization ones.}
\label{fig:ablation}
\end{figure}

Now we conduct controlled experiments to analyse the importance of each component in our model. \tabref{tab:ablation} reports the performance of each model by removing one component from the full model. Again, we argue that these commonly used evaluation metrics does not necessarily reflect the quality of the recovered 3D geometry.
For example, the model with no edge length regularization achieves the best performance across all, however, in fact produces the worst mesh (\figref{fig:ablation}, the last 2nd column). As such, we use qualitative result \figref{fig:ablation} to show the contribution of each component in our system.


\begin{table}[t]
\centering
\setlength{\tabcolsep}{0.95mm}
\renewcommand{\arraystretch}{1.2}
\small
\begin{tabular}{@{}lrrrrrrr@{}}
\hline
Category & -ResNet & -Laplacian & -Unpooling & -Normal & -Edge length & Full model\\
\hline
\hline
F ($\tau$)$\uparrow$ & 55.308 & 60.801 & 60.222 & 58.668 & 60.101 & 59.728 \\
F ($2\tau$)$\uparrow$ & 71.567 & 75.202 & 76.231 & 74.276 & 76.053 & 74.191 \\
CD$\downarrow$ & 0.644 & 0.596 & 0.561 & 0.598 & 0.552 & 0.591 \\
EMD$\downarrow$ & 1.583 & 1.350 & 1.656 & 1.445 & 1.479 & 1.380 \\
\hline
\end{tabular}
\caption{Ablation study that evaluates the contribution of different ideas to the performance of the presented model. The table reports all 4 measurements. For F-score, larger is better. For CD and EMD, small is better. }
\label{tab:ablation}
\end{table}

\subsubsection{Graph Unpooling}
We first remove the graph unpooling layers, and thus each block has the same number of vertices as in the last block of our full model. It is observed that the deformation makes mistake easier at beginning, which cannot be fixed later on. Consequently, there are some obvious artifacts in some parts of the objects.

\subsubsection{G-ResNet}
We then remove the shortcut connections in G-ResNet, and make it regular GCN. As can be seen from \tabref{tab:ablation}, there is a huge performance gap in all four measurement metrics, which means the failure of optimizing Chamfer distance. The main reason is the degradation problem observed in the very deep 2D convolutional neural network. Such problem leads to a higher training error (and thus higher testing error) when adding more layers to a suitably deep model \cite{HeZRS16}. Essetially, our network has 42 graph convolutional layers. Thus, this phenomenon has also been observed in our very deep graph neural network experiment.

\subsubsection{Loss terms}
We evaluate the function of each additional terms besides the Chamfer loss. As can be seen in \figref{fig:ablation}, removing normal loss severely impairs the surface smoothness and local details, e.g. seat back; removing Laplacian term causes intersecting geometry because the local topology changes, e.g. the hand held of the chair; removing edge length term causes flying vertices and surfaces, which completely ruins the surface characteristics.
These results demonstrate that all the components presented in this work contribute to the final performance.

\begin{figure}[tbhp]
\centering
\includegraphics[width=1\linewidth]{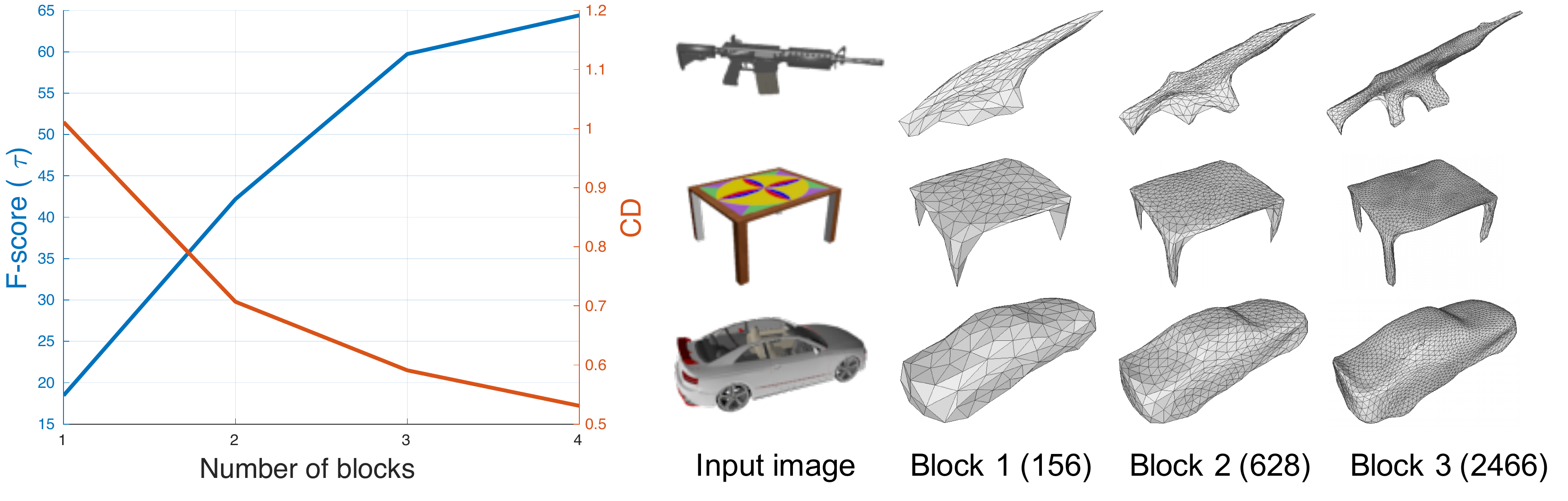}
\caption{Left: Effect of number of blocks. Each curve shows the mean F-score ($\tau$) and CD for different number of blocks. Right: Sample examples showing the output after each block.}
\label{fig:num_block}
\end{figure} 

\subsubsection{Number of Deformation Blocks}
We now analyze the effects of the number of blocks. Figure \figref{fig:num_block} (left) shows the mean F-score($\tau$) and CD with regard to the number of blocks. The results indicate that increasing the number of blocks helps, but the benefit is getting saturated with more blocks, e.g. from 3 to 4. In our experiment, we found that 4 blocks results in too many vertices and edges, which slow down our approach dramatically even though it provides better accuracy on evaluation metrics. Therefore, we use 3 blocks in all our experiment for the best balance of performance and efficiency. 
\figref{fig:num_block} (right) shows the output of our model after each deformation block. Notice how mesh is densified with more vertices and new details are added.

\subsection{Reconstructing Real-World images}
\begin{figure}[t]
\centering
\includegraphics[width=1\linewidth]{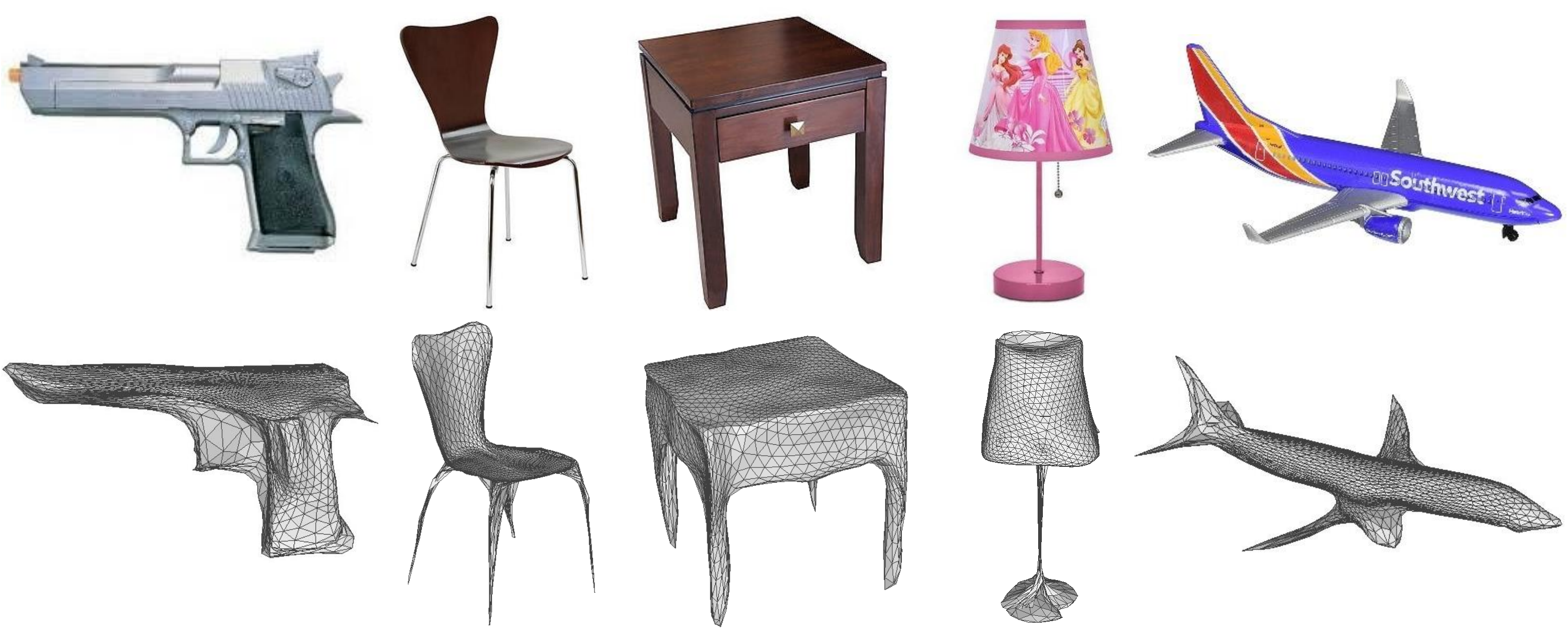}
\caption{Qualitative results of real-world images from the Online Products dataset and Internet.}
\label{fig:real}
\end{figure}

Following Choy et. al. \cite{ChoyXGCS16}, we test our network on the Online Products dataset and Internet images for qualitative evaluation on real images. We use the model trained from ShapeNet dataset and directly run on real images without finetuning, and show results in \figref{fig:real}. As can be seen, our model trained on synthetic data generalizes well to the real-world images across various categories.

\section{Conclusion}
\label{sec:con}
We have presented an approach to extract 3D triangular meshes from singe images. We exploit the key advantages the mesh presentation can bring to us, and the key issues required to solve for success. The former includes surface normal constraints and information propagation along edges; the latter includes perceptual features extracted from images as a guidance. We carefully design our network structure and propose a very deep cascaded graph convolutional neural network with ``shortcut'' connections. Meshes are progressively refined by our network trained end-to-end with the chamfer loss and normal loss.
Our results are significantly better than the previous state-of-the-art using other shape representations such as 3D volume or 3D point cloud. Thus, we believe mesh representation is the next big thing in this direction, and
we hope that the key components discovered in our work can support follow-up works that will further advance direct 3D mesh reconstruction from single images.

\subsubsection{Future work}
Our method only produces meshes with the same topology as the initial mesh. In the future, we will extend our approach to more general cases, such as scene level reconstruction, and learn from multiple images for multi-view reconstruction.

\subsubsection{Acknowledgements}
This work was supported by two projects from NSFC (\#61622204 and \#61572134), two projects from STCSM (\#16JC1420401 and \#16QA1400500), Eastern Scholar (TP2017006),
and The Thousand Talents Plan of China (for young professionals, D1410009).

%
%
%
\bibliographystyle{splncs04}
\bibliography{egbib}
\end{document}



\title{Pixel2Mesh: Generating 3D Mesh Models \\from Single RGB Images \\(Supplementary Material)} 

\titlerunning{Pixel2Mesh}

\authorrunning{ECCV-18 submission ID \ECCV18SubNumber}

\author{Nanyang Wang$^{1}$\thanks{∗ indicates equal contributions.}, ~~Yinda Zhang$^{2\,\star}$, ~~Zhuwen Li$^{3\,\star}$, \\Yanwei Fu$^{1}$, ~~Wei Liu$^{4}$, ~~Yu-Gang Jiang$^{1}$}
\authorrunning{N. Wang$^\star$, Y. Zhang$^\star$, Z. Li$^\star$, Y. Fu, W, Liu, Y. Jiang}

\institute{\textsuperscript{1}Fudan University\hspace{3mm}\textsuperscript{2}Princeton University\hspace{3mm}\textsuperscript{3}Intel Labs\hspace{3mm}\textsuperscript{4}Tencent AI Lab}

\maketitle

In this supplementary material, we provide additional implementation details, analysis of the sensitivity of our results with the initial mesh, comparsions to other related methods, additional metrics for the ablation study, different views of the reconstructed mesh, and more qualitative results on ShapeNet images and real-world images.

\section{Implementation Details}
\subsection{Camera projection}
The perceptual feature pooling layer projects a 3D vertex onto the image plane and pools features from the image feature pathway.
For a 3D vertex with coordinate $(X,Y,Z)$, its 2D projection in image is:
\begin{align}
    x &= \frac{X}{Z}*f_x + c_x, \nonumber\\
    y &= \frac{Y}{Z}*f_y + c_y,
    \label{equ:projection}
\end{align}
where $f_x,f_y$ are focal lengths along horizontal and vertical image axis, and $(c_x,c_y)$ is the center of projection of the camera.
These parameters can be readily obtained by intrinsic camera calibration once for a camera.

\subsection{Bilinear feature pooling}

Suppose the coordinate of the projection is $(x,y)$. The feature is pooled using bilinear interpolation \cite{bilinear}:
\begin{align}
    f = \frac{(x_2-x)(y_2-y)}{(x_2-x_1)(y_2-y_1)}f(x_1,y_1) +\frac{(x-x_1)(y_2-y)}{(x_2-x_1)(y_2-y_1)}f(x_2,y_1) \nonumber\\ +\frac{(x_2-x)(y-y_1)}{(x_2-x_1)(y_2-y_1)}f(x_1,y_2) +\frac{(x-x_1)(y-y_1)}{(x_2-x_1)(y_2-y_1)}f(x_2,y_2),
\end{align}
where $(x_1,y_1)$, $(x_1,y_2)$, $(x_2,y_1)$ and $(x_2,y_2)$ are the integral coordinates of the pixel square where the projection reside in.



\section{Comparison with Octree based Voxel Generation Method}
This section provides a comparison with a Octree based voxel Reconstruction method \cite{TatarchenkoDB17}, which produces higher resolution voxels. Following Tartarchenko \etal \cite{TatarchenkoDB17}, we train on Shapenet-car and compare to it as shown in Tab. \ref{tab:octree_car}. Our model consistently outperform the octree based approach in all the evaluation metrics.
\begin{table}[h]
\centering
\setlength{\tabcolsep}{1.5mm}
\begin{tabular}{c|cccc}
    \hline
    Method. & {F-score($\tau$)$\uparrow$} & {F-score(2$\tau$)$\uparrow$} & {CD$\downarrow$} & {EMD$\downarrow$}\\
    \hline
    Tartarchenko et al & 65.335 & 79.733 & 0.361 & 1.273 \\
    Ours & 72.128 & 87.247 & 0.236 & 1.220\\
    \hline
\end{tabular}
\caption{Comparsion to Tartarchenko et al.($128^3$) on ShapeNet-cars.}
\label{tab:octree_car}
\end{table}
\section{Hausdorff Distance for the Ablation Study}
Hausdorff distance  \cite{haufdist} between meshes measures how far two meshes are from each other, 
we evaluate the ablation study with Hausdorff distance in Tab. \ref{tab:ablation_hd}. The full model outperforms all except the one without Laplacian. Again, we think Laplacian term regulates deformation and thus benefits the surface smoothness and continuity, which is hard to be reflected in Hausdorff distance.

\begin{table}[h]
\centering
\setlength{\tabcolsep}{0.95mm}
\begin{tabular}{c|cccccc}
\hline
 & -ResNet & -Laplacian & -Unpooling & -Normal & -Edge length & Full model\\
\hline
HD$\downarrow$ & 0.644 & 0.600 & 0.616 & 0.623 & 0.643 & 0.603 \\
\hline
\end{tabular}
\caption{Results for the ablation study on hausdorff distence.}
\label{tab:ablation_hd}
\end{table}
\section{Other Views of the Generated Mesh} 
We provide mesh visualizations from other viewpoints, please see \figref{fig:different_view} for two examples. As can be seen, our model also recovers invisible parts of the 3D mesh.
\begin{figure}[t]
\centering
\includegraphics[width=1\linewidth]{figures/new_fig/rebuttal_figure.pdf}
\vspace{-20pt}
\caption{Mesh visualization from a different view.}
\label{fig:different_view}
\end{figure}

\section{Sensitivity to the Initial Meshes}
We test our approach with different initial meshes. Specifically, we have tried sphere, noise ellipsoid (EllipsoidN) and ellipsoid in vertical (EllipsoidV) in addition to the original ellipsoid in horizontal (EllipsoidH) as shown in \figref{fig:quali_init}. With each different shape, we fine-tune the model with 30,000 iterations and report quantitative and qualitative results in the \figref{fig:quali_init} and \tabref{tab:quant_rel}. As can be seen, our method is not sensitive to the initial shape.


\begin{table}[t]
\centering
\setlength{\tabcolsep}{0.55mm}
\renewcommand{\arraystretch}{1.2}
\small
\begin{tabular}{@{}lccccccccc@{}}
\hline
& \multicolumn{4}{@{}c@{}}{F-Score($\tau$)} & \multicolumn{4}{@{}c@{}}{CD} \\
  \cmidrule(lr){2-5} \cmidrule(lr){6-9}
Category  & Sphere & EllipsoidN & EllipsoidV & EllipsoidH & Sphere & EllipsoidN & EllipsoidV & EllipsoidH\\
\hline
\hline
  plane & 69.11 & 70.52 & 68.68 & 71.12 & 0.482 & 0.463 & 0.491 & 0.477\\
  bench & 55.02 & 57.20 & 54.78 & 57.57 & 0.700 & 0.661 & 0.744 & 0.624\\
  cabinet & 57.35 & 60.26 & 56.77 & 60.39 & 0.425 & 0.393 & 0.463 & 0.381\\
  car & 64.18 & 67.50 & 62.16 & 67.86 & 0.313 & 0.282 & 0.367 & 0.268\\
  chair & 51.36 & 54.12 & 51.19 & 54.38 & 0.690 & 0.645 & 0.740 & 0.610\\
  monitor & 48.19 & 51.58 & 48.30 & 51.39 & 0.858 & 0.801 & 0.882 & 0.755\\
  lamp & 46.54 & 47.70 & 46.63 & 48.15 & 1.372 & 1.355 & 1.454 & 1.295\\
  speaker & 46.64 & 49.59 & 47.29 & 48.84 & 0.798 & 0.748 & 0.835 & 0.739\\
  firearm & 70.69 & 71.39 & 71.52 & 73.20 & 0.468 & 0.462 & 0.465 & 0.453\\
  couch & 48.87 & 52.26 & 48.62 & 51.90 & 0.587 & 0.527 & 0.605 & 0.490\\
  table & 63.34 & 65.37 & 63.21 & 66.30 & 0.547 & 0.505 & 0.590 & 0.498\\
  cellphone & 66.20 & 68.56 & 64.23 & 70.24 & 0.489 & 0.455 & 0.527 & 0.421\\
  watercraft & 52.53 & 54.21 & 52.07 & 55.12 & 0.737 & 0.707 & 0.758 & 0.670\\
\hline
  mean & 56.93 & 59.25 & 56.58 & 59.72 & 0.651 & 0.616 & 0.686 & 0.591\\
\hline
\end{tabular}
\vspace{3mm}
\caption{F-Score ($\tau$) and CD measurements with different initial shapes. ElliposidN, ElliposidV and ElliposidH represent ellipsoid with noise, in vertical and in horizontal respectively. In the main text, we adopt the ellipsoid in horizontal.}
\label{tab:quant_rel}
\end{table}

\section{More Qualitative Results}
\subsection{ShapeNet dataset}
We show more qualitative results on rendered images from the ShapeNet dataset in Fig. \ref{fig:quali_shapenet}. The first column shows the color images; the 2nd column shows the volume results from Choy \etal \cite{ChoyXGCS16} and the mesh converted using Marching Cube \cite{LorensenC87}; the 3rd column shows the point cloud from Fan \etal \cite{FanSG16} and the mesh converted using Ball Pivoting \cite{BernardiniMRST99}; the 4th column shows the results of Neural 3D Mesh Renderer \cite{KatoUH2018} and the last column shows our results.
Notice how our method produces both smooth surface and the sharp details compared to other methods.

\begin{figure}[tbhp]
\centering
\includegraphics[width=1\linewidth]{figures/new_fig/more_shapenet.pdf}
\caption{Qualitative Results on ShapeNet Dataset. We compare with 3DR2N2 \cite{ChoyXGCS16}, PSG \cite{FanSG16}, and N3MR \cite{KatoUH2018}. Our method produces results with smooth surface and sharp details.}
\label{fig:quali_shapenet}
\end{figure}

\begin{figure}[tbhp]
\centering
\includegraphics[width=1\linewidth]{figures/new_fig/initial.pdf}
\caption{Qualitative Results of Using Different Initial Mesh. The left column shows the input images. The 2-5 column shows the results using different initial mesh. Our method is not sensitive to the initial 3D mesh.}
\label{fig:quali_init}
\end{figure}

\begin{figure}[tbhp]
\centering
\includegraphics[width=1\linewidth]{figures/new_fig/more_real.pdf}
\caption{Qualitative Results on Real-world Images. Each row shows three examples of our results on real-world images.}
\label{fig:quali_real}
\end{figure}

\subsection{Real-world images}
We show more results on real-world images in Fig. \ref{fig:quali_real}. As can be seen, our method generalizes well to real-world images.

\bibliographystyle{splncs04}
\bibliography{egbib}